# Automatic Number Plate Recognition (ANPR) with YOLOv3-CNN

Rajdeep Adak, Abhishek Kumbhar, Rajas Pathare, Sagar Gowda
K. J. Somaiya College of Engineering, University of Mumbai

## Abstract

Automatic Number Plate Recognition (ANPR) is a technique designed to read vehicle number plates without human intervention using high speed image capture with supporting illumination, detection of characters within the images provided, verification of the character sequences as being those from a vehicle numberplate, character recognition to convert image to text; so ending up with a set of metadata that identifies an image containing a vehicle numberplate and the associated decoded text of that plate. Most members of the public will be aware that ANPR is used by police forces to track criminal behavior on road. However, ANPR is used in a variety of other ways to support the security and safety of the public as well as supports the way we interact with transportation and vehicle based infrastructure. The software aspect of the system runs on standard home computer hardware and can be linked to other applications or databases. It first uses a series of image manipulation techniques to detect, normalize and enhance the image of the number plate, and then optical character recognition (OCR) to extract the alphanumeric of the license plate. ANPR systems are generally deployed in one of two basic approaches: one allows for the entire process to be performed at the lane location in real-time, and the other transmits all the images from many lanes to a remote computer location and performs the OCR process there at some later point in time. This information can easily be transmitted to a remote computer for further processing if necessary, or stored at the lane for later retrieval. In the other arrangement, there are typically large numbers of PCs used in a server farm to handle high workloads.

Often in such systems, there is a requirement to forward images to the remote server, and this can require larger bandwidth transmission media. At the front-end of any ANPR system is the imaging hardware which captures the image of the license plates. The initial image capture forms a critically important part of the ANPR system which, in accordance to the garbage in, garbage out principle of computing, will often determine the overall performance. License plate capture is typically performed by specialized cameras designed specifically for the task.

ANPR was first implemented in 1990s and since then it has come a long way to become more efficient and cost effective. These features have enabled large scale implementation in law enforcements in a number of countries. However, ANPR still faces some challenges. Vehicle owners have used a variety of techniques in an attempt to evade ANPR systems and road-rule enforcement cameras in general. One method increases the reflective properties of the lettering and makes it more unlikely for the system to locate the plate or produce a high enough level of contrast to be able to read it. This is typically done by using a plate cover or a spray, though claims regarding the effectiveness of the latter are disputed. The introduction of ANPR systems has led to fears of misidentification. Many ANPR systems claim accuracy when trained to match plates from a single jurisdiction or region, but can fail when trying to recognize plates from other jurisdictions due to variations in format, font, color, layout, and other plate features.

This documentation presents implementation of a deep learning based ANPR technique. We test the program based on our analysis on real-life traffic videos and store numberplate text and image into a directory. Any number plate can be searched from this directory as the file name is the predicted text from the implemented ANPR algorithm. Results demonstrate the accuracy of the ANPR technique on various scenarios. A detailed description of pre-processing techniques and alphabet/number extraction algorithms are presented. Multiple alternative methods have also been compared to determine the optimal solution. We then describe our program architecture and its various components and tuning parameters which can be manually changed to obtain desired results. A performance review describes robustness, utility, cost, complexity and ease of deployment. We have made our datasets of images and videos, theoretical analysis, results of experiments, program architecture and code available for further exploration and contribution. The source code of the ANPR system is open sourced and can be accessed on GitHub. We regularly update new commits to the GitHub repository.

Further work such as multiple predicted outcomes and corresponding confidence level calculation, storing and retrieving data efficiently over a database are being developed.

## Keywords

Automatic Number Plate Recognition (ANPR), Contour Detection, Segmentation, Convolutional Neural Networks (CNN), YOLO, Darknet, CUDA, Bilateral Filter, Canny Edge Detection, Blob Removal, Adaptive Thresholding, Otsu's Thresholding. Sliding Concentric Window (SCW), Fuzzy-based Algorithm, Probabilistic Neural Network (PNN), Character Segmentation using Image Scissoring, Character Segmentation using Blob removal, Artificial Neural Network (ANN), Template Matching, Self-organized maps.

# Contents

# List of Figures

# List of Tables

# Chapter 1 Introduction

> This chapter introduces the concept of ANPR and how it emerged into wide scale usable technology. Our approach is based specifically on Indian number plates and hence we have proposed a methodology accordingly. The project focuses on the software part of a general ANPR system and proposes an efficient alternative to many other approaches previously implemented or proposed. The report is organized in which the program has been written. Hence it is in a step-by-step format. We recommend readers to read the Introduction and Literature Review as they form the foundation of our methodology.

## 1.1 Background

Unlike other countries, India, with its one billion people population, has a unique set of needs for ANPR. The main use of ANPR is in highway monitoring, parking management, and neighborhood law enforcement security.

In India there is one death in every four minutes with most of them occurring due to overspeeding. ANPR is used to monitor the vehicles' average speed and can identify the vehicles that exceed the speed limit. In this case, a fine ticket can be automatically generated by calculating the distance between two cameras. This helps to maintain law and order which, in turn, can minimize the number of road casualties.

ANPR provides the best solution for providing parking management. Vehicles with registered plates can automatically enter into parking areas while non-registered vehicles will be charged by time of check in and check out. Number plates of the car can be directly linked with owner mobile phone and parking tickets could be paid without any extra effort directly from the user's account against the ticket number generated. ANPR can support a cloud-based system pre-book and pre-pay platform for parking.

In India 200,000 cars are stolen per year. This number can lessen if proper steps are taken and ANPR system is used to track cars so that if vehicles are stolen, law enforcement will be able to identify when, where and the route taken by a stolen vehicle. This can help bring justice swiftly to such a vast nation.

## 1.2 Motivation

In countries where ANPR systems have been implemented required availability of funds and human labor to cover streets and highways with specialized cameras. Some of the cameras perform in-lane processing and send only the text to a remote server to store.
Ordinary CCTV cameras in India record and store the entire video feed in a remote storage. Hence we have designed a program which can utilize the video feed previously recorded by a camera to detect and store the number plate in text format in the local machine where the detection was performed.

## 1.3 Scope of the project

The project focuses on the implementation of a deep learning based technique to detect characters in a number plate. The algorithm can run on a local machine consisting of required libraries and software already installed. The algorithm is also scalable to micro- processing units other than ordinary computers that have an OS such as Windows/Linux.

## 1.4 Organization of the report

The first chapter explains the status of ANPR in India. In India even though CCTV cameras have now become common, they are yet to be integrated with an ANPR system. Number plates in India are very diverse. Some of them are even handwritten and hence standard OCR techniques often fail to detect such number plates. Chapter 2 provides various techniques which have been proposed in India as well as in other countries to produce credible outputs. Chapter 3 provides the program architecture and flow chart.

Chapter 4 provides a detailed description of our methodology, in-depth analysis, tuning parameters and comparison of various processing techniques which can be used for the purpose. Chapter 5 consists of our conclusions, a review of its performance on various environmental factors and further scope of work.

## Chapter 2 Literature survey

> This chapter provides previously done work in the field of ANPR. It mentions important features of various techniques utilized to obtain accurate output. We then contemplate our methodology based on the congregation of those methods which produced suitable outputs.

The methods discussed in the following sections are common methods for plate detection. As most of the methods discussed in these literatures use more than one approach, it is not possible to do category wise discussion. Different number plate segmentation algorithms are discussed below.

### 2.1   Sliding Concentric Window (SCW)

In [1], for faster detection of region of interest (ROI) a technique called sliding concentric window (SCW) is developed. It is a two-step method contains two concentric windows moving from upper left corner of the image. Then statistical measurements in both windows were calculated based on the segmentation rule which says that if the ratio of the mean or median in the two windows exceeds a threshold, which is set by the, then the central pixel of the windows is considered to belong to an ROI. The two windows stop sliding after the whole image is scanned. The threshold value can be decided based on trial and error basis. The connected component analysis is also used to have overall success rate of 96%.

## 2.2    Fuzzy-based Algorithm

In [2], fuzzy-based algorithm is applied. To extract license plate region a four step method is implemented. In the first step noise is eliminated from the input image. Edge detection is used in second step of find rectangle area of candidate region. In the third step, based on size, histogram and other information invalid rectangle areas are discarded. In the last step geometric rectification is used to obtain license plate candidate region. As these steps need some addition processing, authors used fuzzy-based algorithm containing several steps to extract license plate with more accuracy. The system was developed on TI DM642 600 MHz/32 MB RAM with C language under CCStudio V3.1 environment. The overall average processing time was ~418.81ms.

## 2.3    Probabilistic Neural Network (PNN)

A probabilistic neural network (PNN) based approach is presented in [3]. The PNN algorithm works on gray scale image. Bottom-Hat filtering is used to enhance the potential plate regions. To separate the object of interest from background a Thresholding is employed for binarization of the gray level image. Because of varying lighting conditions, brightness levels may vary and some adaptation is necessary. To perform it Otsu's Thresholding technique is used as it is adaptive in nature. Each segment of the binary image is labelled according to color of each segment to enable classification. The plate extraction is done calculating the Column Sum Vector (CSV) and its local minima. The algorithm was executed on Intel® Core™2 Duo Processor CPU P8400 (2.26GHz, 2267 MHz). The plate recognition processing time was 0.1s.

## 2.4     Character Segmentation using Image Scissoring

Prathamesh Kulkarni et al. [4] conclude that blob coloring and peak-to-valley methods are not suitable for Indian number plate. The authors proposed image scissoring algorithm in which a number plate is vertically scanned and scissored at the row where there is no white pixel and this information is stored in the matrix. In case of more than one matrix, a false matrix is discarded based on the formula given in this paper. Same process is repeated for horizontal direction by taking width as a threshold.

## 2.5     Character Segmentation using Blob removal

H.Erdinc Kocer [5] used contrast extension, median filtering and blob coloring methods for character segmentation. Contrast extension is used to make image sharp. As per H.Erdinc Kocer the histogram equalization is a popular technique to improve the appearance of a poor contrasted image. In median filtering unwanted noisy regions are removed. Blob coloring method is applied to binary image to detect closed and contact less regions. In this method, an L shaped template is used to scan image from left to right and top to bottom. This scanning process is used to determine the independent regions by obtaining the connections into four directions from zero valued background. The four directional blob coloring algorithm is applied to the binary coding license plate image for extracting the characters. At the end of this process the numbers are segmented in the size of 28 X 35 and letters are segmented in the size of 30 X 40.

Character recognition helps in identifying and converting image text into editable text. As most of the number plate recognition algorithms use single method for character recognition. In this section, each method is explained.

## 2.6 Character Recognition using Artificial Neural Network (ANN)

Artificial Neural Network (ANN) sometimes known as neural network is a mathematical term, which contains interconnected artificial neurons. In [3] two-layer probabilistic neural network with the topology of 180-180-36. The character recognition process was performed in 128ms.

## 2.7 Character Recognition using Template Matching

Template matching is useful for recognition of fixed sized characters. It can be also used for detection of objects generally in face detection and medical image processing. It is further divided in two parts: feature based matching and template based matching.

Feature based approach is useful when template image has strong features otherwise template based approach can be useful. In [4] statistical feature extraction method is applied for achieving 85% of character recognition rate. In [6], several features and extracted and salient is computed based on training characters. A linear normalization algorithm is used to adjust all characters with uniform size. The recognition rate of 95.7% is achieved among 1176 images. An SVM based approach is used for feature extraction of Chinese, Kana and English, Numeric characters. The authors achieved success rate of 99.5%, 98.6%, and 97.8% for numerals, Kana, and address recognition respectively.

## 2.8 Other methods

In some algorithms character recognition is done by the available Optical Character Recognition (OCR) tool. There are numerous software available for OCR processing. One of the open source OCR tools with multilingual supported is Tesseract which is maintained by Google and available at [7]. It is used in [8] for character recognition. The author modified it to achieve 98.7% of character recognition rate.

The method proposed in [9] consists of three steps: character categorization, topological sorting and self-organizing (SO) recognition. Character categorization is used to classify character as alphabet or number. In second step topological features of input character are computed and compared with already stored character templates. Compatible templates will form a test set, in which the character template that best matches the input character is determined. The template test is performed by a SO character recognition method Self-organized neural network is based on Kohonen's self-organized feature maps to handle noisy, broken, or incomplete characters. To differentiate the similar characters from character pairs such as (8, B) and (O, D) the authors predefined an ambiguity set containing the characters 0, 8, B and D.

After analyzing the methods used in each of the above ANPR techniques a neural network based character recognition approach is decided to be an optimal alternative to recognize individual alphabets and numbers of the English alphabet and number system. The images of an environment consisting of vehicles with visible number plates are first filtered to extract the number plate and then further filtering such as blob removal and edge detection are applied to produce a transform of the number plate. The detailed mechanism is explained in the next chapter.

# Chapter 3 Project Design

This chapter explains our approach to build the ANPR system. The techniques mentioned below allow segregation of images of number plates from the image of an environment consisting of a vehicle. Certain terms related to Image-processing and Deep Learning are required to understand how the system works. These terms have a brief description in the Concepts section. Readers are recommended to go through them before proceeding to the next chapter. The program architecture section shows the flow of the code and how it converts images of characters to text.

## 3.1  Approach

Our $1^{st}$ region of interest (ROI) is the numberplate itself. To segregate the number plate from the rest of the image we utilize YOLOv3[10]. YOLO, which stands for 'You Only Look Once', is a state-of-the-art Object Detection Algorithm. In comparison to recognition algorithms, a detection algorithm does not only predict class labels but detects locations of objects as well. So, it not only classifies the image into a category, but it can also detect multiple Objects within an image. This Algorithm doesn't depend on multiple neural networks. It applies a single neural network to the Full Image. This network divides the image into regions and predicts bounding boxes and probabilities for each region. These bounding boxes are weighted by the predicted probabilities. Our code for object detection will use weights trained from custom dataset of 'Number Plates' collected and labelled solely by the team using Darknet Yolo v3. Darknet is an open source neural network framework written in C and CUDA. It is fast, easy to install, and supports CPU and GPU computation. The weights used in Darknet Yolo v3 are converted into '.h5' file, weights used by Keras. An h5 file is a data file saved in the Hierarchical Data Format (HDF).

This converted weights or Keras Model is then used to predict the bounding boxes and class probabilities using Keras library. Once the number plate is detected, it is cropped from the image and is passed on to detect the 2$^{nd}$ Region of Interest (ROI) i.e. individual characters. Various Image Processing algorithms are used to perform filtering of content which isn't a character or a number. The resultant image consists figures which include all characters and numbers present in the numberplate along with those which are probable characters and number plates.

## 3.2   Concepts

To achieve only characters from an image/video of an entire environment requires implementing the concepts of Edge Detection, Pattern Recognition, Region of Interest Detection, Image Filtering, Contour Detection, Contour Segmentation, and Convolutional Neural Networks. The above concepts are briefly explained below and have been applied methodically to obtain optimal results.

**Edge Detection**: Edge detection is an image processing technique for finding the boundaries of objects within images. It works by detecting discontinuities in brightness.

**Pattern Recognition**: Pattern recognition is the ability to detect arrangements of characteristics or data that yield information about a given system or data set. In a technological context, a pattern might be recurring sequences of data over time that can be used to predict trends, particular configurations of features in images that identify objects, frequent combinations of words and phrases, or particular clusters of behavior on a network etc.

**Region of Interest (ROI)**: A region of interest (often abbreviated ROI), are samples within a data set identified for a particular purpose. The concept of a ROI is commonly used in many application areas. In our case, the 1$^{st}$ ROI is the numberplate and 2$^{nd}$ ROI is the characters (A-Z or a-z) and numbers (0-9) in the numberplate.

**Image Filtering**: To remove noise and various unwanted shapes and impurities from the image a number of pre-processing algorithms have been applied. These allow elimination of fake characters which can be wrongly detected.

**Contour Detection**: Contours detection is a process can be explained simply as a curve joining all the continuous points (along with the boundary), having same color or intensity. The contours are a useful tool for shape analysis and object detection and recognition. We utilize this to find characters and numbers as they are continuous figures (this however depends upon the chosen font as some fonts may not keep all characters and numbers as continuous figures).

**Contour Segmentation**: Segmentation is a section of image processing for the separation or segregation of information from the required target region of the image. Once contours are detected they can easily be separated from the rest of the image.

**Convolutional Neural Networks**: A Convolutional Neural Network (ConvNet/CNN) is a Deep Learning algorithm which can take in an input image, assign importance (learnable weights and biases) to various aspects/objects in the image and be able to differentiate one from the other. The pre-processing required in a ConvNet is much lower as compared to other classification algorithms. While in primitive methods filters are hand-engineered, with enough training, ConvNets have the ability to learn these filters/characteristics

## 3.3 Program Architecture

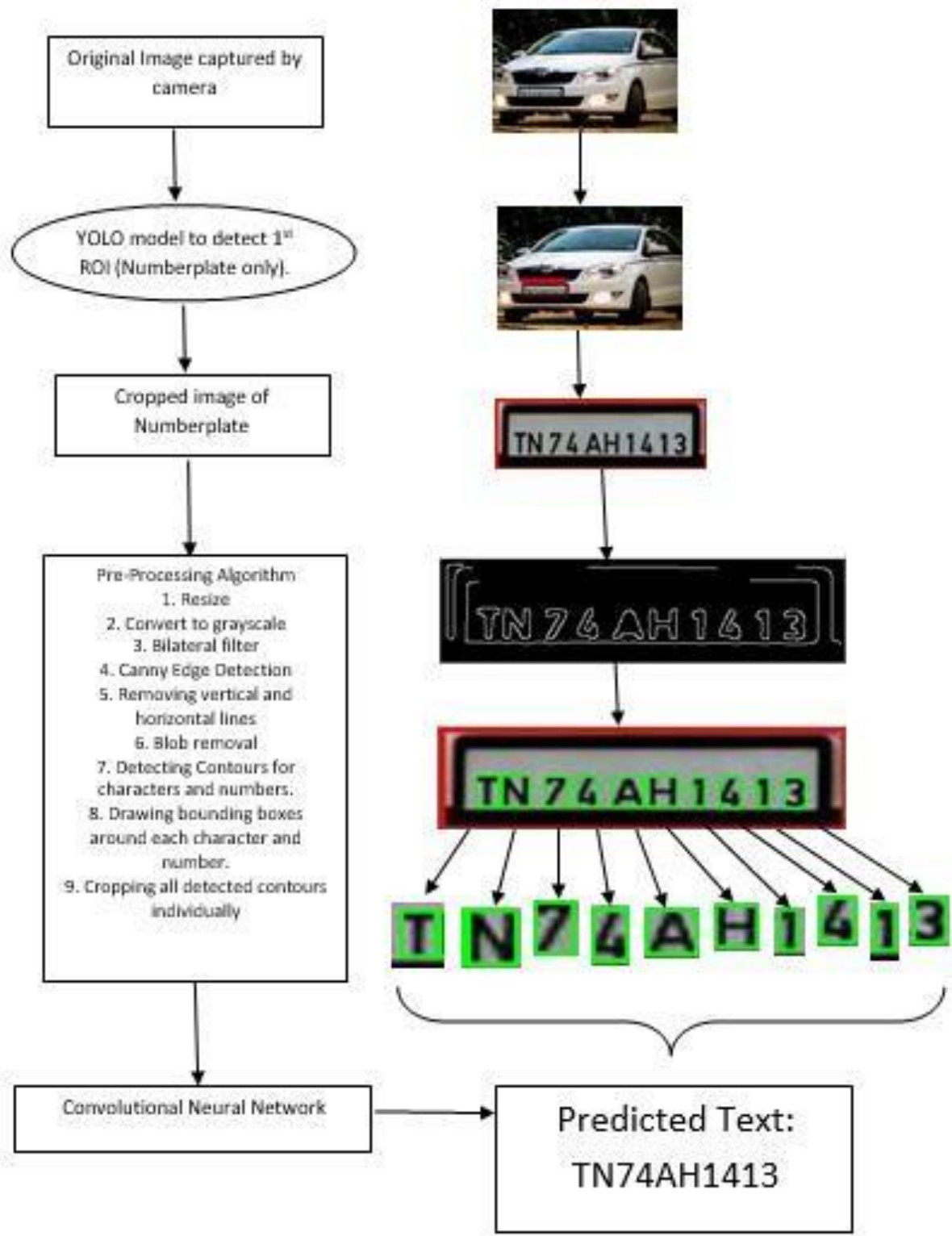

Figure 1. Program logic flow

# Chapter 4 Implementation and Experimentation

> The program was designed and implemented on images affected by various environmental factors. The dataset used to train the deep learning model has several images of characters where clear distinguishability is not readily available. This helps the CNN to develop feature maps for such diverse factors also. The section on Effect of Environmental Factors tabularizes the outcome. Further we have taken inference on the characteristics of our code and explained methods to tune the parameters to obtain better outputs. This can be done manually.

## 4.1 Dataset used

The dataset used to train the CNN has over 2000 images of alphabets and numbers. Each of them have are transformed using image processing algorithms to generate the effect various practical situations such as angular viewing, perspective transform, shadowed images etc.

## 4.2 Effect of Environmental Factors

After training the CNN the program was tested on a large set of images. The overall accuracy obtained is 83 %.

The environmental factors we have aimed to mitigate are:

1. Over Exposed Images.
2. Blurred Images.

3. Images with angular perspective.
4. Randomly chosen custom font.
5. Rotated Images.
6. Shaded Images.
7. Straight Image.
8. Image with a colored Numberplate background.
9. Multiline Image.

The following table consists of 10 images each of which have an environmental factor. Their character accuracy, predicted number plate label and processing time have been shown.

| Sr. No. | Environmental Factor | Predicted label | True label | character accuracy | processing time (sec) |
|---|---|---|---|---|---|
| 1 | Angular perspective | DL3CBD5092 | DL3CBD5092 | 100% | 0.308 |
| 2 | Blurred | 73H12AF5032 | MH12AF5032 | 90% | 0.197 |
| 3 | Custom Font 1 | KL59T997 | KL59T997 | 100% | 0.44 |
| 4 | Custom Font 2 | HR658795 | HR16S8179 | 60% | 0.346 |
| 5 | Multiline | MH14GN9239 | EFN9239 | 50% | 0.4 |
| 6 | Over exposed | MH02EP6969 | MH02EP6969 | 100% | 0.464 |
| 7 | Rotated | AP05BL6339 | AP05BL6339 | 100% | 0.19 |
| 8 | Shaded | GJH06C122 | JH06C1122 | 77% | 0.59 |
| 9 | Straight front | MH20DV2362 | MH20DV2362 | 100% | 0.75 |
| 10 | Colored background | DL1RTA2179 | DL1RTA2179 | 100% | 0.3 |

Table 1. Environmental factor vs. accuracy and processing time

Images chosen for this purpose have been shown below:

Figure 2. Images of all cars with environmental factors

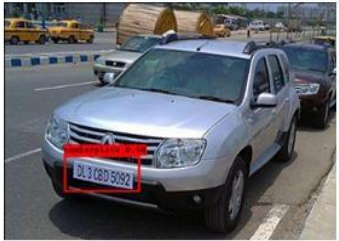
Angular perspective

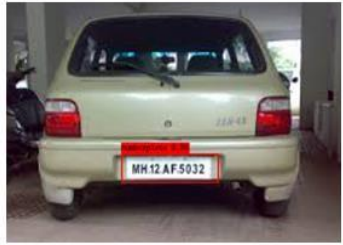
Blurred

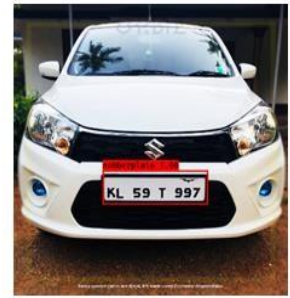
Custom Font

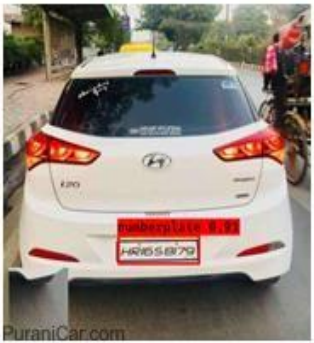
Custom font

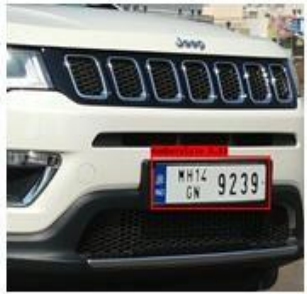
Multiline

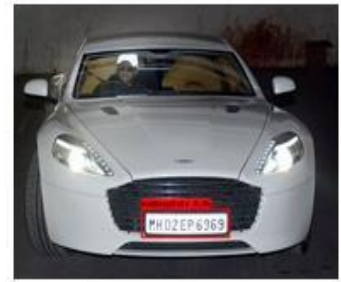
Over Exposed

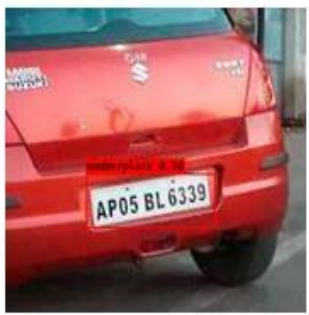
Rotated

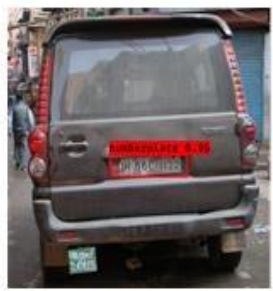
Shaded

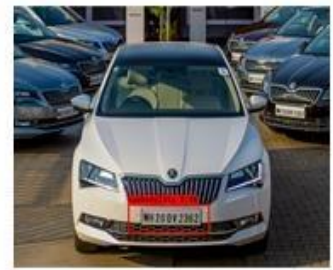
Straight front

**Department of Electronics and Telecommunication Engineering Semester VI 2017-21 Batch** Page No 15

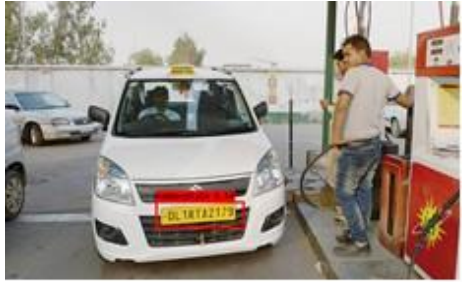

Coloured Background Numberplate

**Observations**

| | |
|---|---|
| Average accuracy obtained on a set of 10 images with environmental factors = 87.7% | |
| Average accuracy on a set of 10 straight front images (with no environmental factors) = **95 %** | |
| Difference in accuracy caused by presence of environmental factors = 95% – 87.7% = **7.3 %** | |
| Difference in overall accuracy = 95% - 83% = **12%** | |
| Average processing time = **0.3985 seconds**. | |

## 4.3 Inference

From the observations it is inferred that on straight front images the ANPR program works with an accuracy of up to 95%. Due to presence of environmental factors a deficit of 7.3% - 12% is observed. Hence the ANPR program will produce unsatisfactory results up to 12 % of all cases. Upon analysis of the reason behind 12 % inaccuracy we have deduced 2 major causes:

**1.    Improper contours produced by the pre-processing steps**.

The contour detection works on the basis of finding continuous pixels of same intensity or color. Certain environmental factors affect the contour detection as characters or numbers may not be distinctly segregated. This can appear in case when the gap between two characters is too small, presence of a mounting nail, blurred image, shadow causing lowering of intensity etc. Upon changing the parameters utilized in the bilateral filter and canny edge detection functions we have observed a better result.

However, the program is a generalized code and hence only the meticulously chosen parameter values have been written. Besides Canny edge detection algorithm, Adaptive Thresholding algorithm and Otsu's Thresholding have also shown good results. Some images with altered parameter values have been shown below that mitigate the environmental factor affecting the accuracy.

**2.   Ambiguous characters and numbers improperly classified.**

Certain characters and numbers such as the letter G and C, number 6 can be wrongly detected as they have similar feature maps. It depends on how well the CNN is trained to efficiently distinguish between such ambiguous characters.

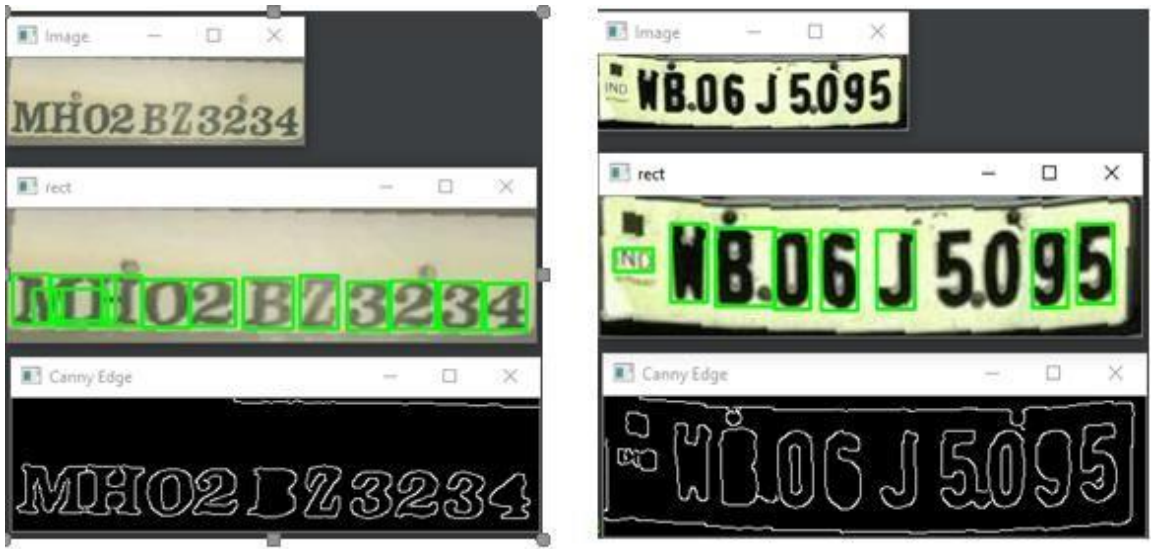

Here the letters M and H appear as continuous contours in output of pre-processing steps. Hence the character is incorrectly determined

The numbers 5 and 0 have a dot in between which seems to be a mounting nail. The contour detection fails as they are detected as a single character but out of the bounds of the aspect ratio of a single character. Hence not highlighted.

Figure 3. Ambiguous characters and failed contouring

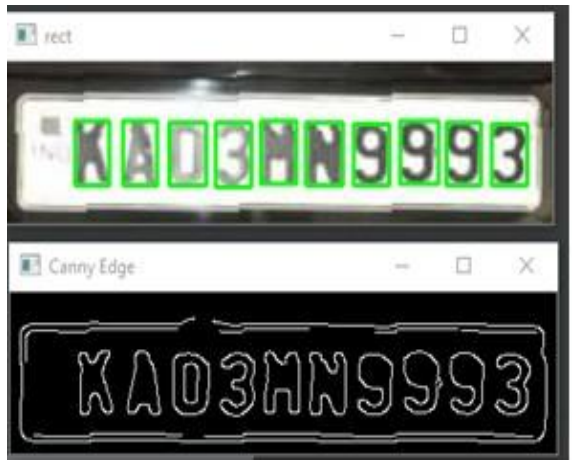

Here even though all the characters are well separated and detected effectively by the pre-processing algorithm the output is however = KA03HN9993. The letter M and H have an ambiguity and it is clearly observed in the image.

Figure 4. Ambiguous characters but correct output

## 4.4   Tuning Parameters

| Paramter | Method/function using the parameter | default | effect |
|---|---|---|---|
| Horizontal Line size tuple | cv2.getStructuringElement | (10, 1) | Change to increase or decrease size of removable horizontal lines |
| Horizontal Line morphology iteration | cv2.morphologyEx | iterations=8 | Change to increase or decrease the number of iterations performed to remove horizontal lines |
| Vertical Line size tuple | cv2.getStructuringElement | (1, 20) | Change to increase or decrease size of removable Vertical lines |
| Vertical Line morphology iteration | cv2.morphologyEx | iterations=8 | Change to increase or decrease the number of iterations performed to remove Veritcal lines |
| blob size | morphology.remove_small_objects | min_size=50 | Change to increase or decrease size of vlob to be removed |
| contour marking | cv2.findContours | cv2.RETR_EXTERNAL | Change parameter name to detect internal and external contours |
| area | user defined | None | min. and max. area of character contour |
| perimeter | user defined | None | min. and max. perimeter of character contour |
| aspect ratio | user defined | None | min. and max aspect ratio of character contour |
| width | user defined | None | min. and max. width of character contour |
| height | user defined | None | min. and max. height of character contour |
| Bilateral filter kernel | cv2.bilateralFilter | 9, 70, 70 | Change to increase/decrease blur of surfaces without affecting edges |
| canny edge maxima and minima | cv2.Canny | 30, 130 | Change to increase of decrease min and max edge gradient |

Table 2. Tuning Parameters.

Other factors that can affect the accuracy of the algorithm are hardware inadequacies such as low shutter speed, low fps, low resolution, improperly positioned camera, rain, polluted air, dirt etc.

# Chapter 5 Conclusion and Scope for further work

> This chapter concludes the project. We have performed tests on a number of images and defined the effective utility of the ANPR system. As always, such systems are under constant development our team is working consistently to improve the reliability of our algorithm. We have provided links and documentation to the source code. We request readers and developers to issue pull requests to contribute their ideas and valuable advice.

## 5.1 Conclusion

The design ANPR program produces excellent accuracy for straight front images and is hindered up to 12 % due to environmental factors. The same hindrance is observed up to 42 % without the right pre-processing. Hence we have achieved the objective of reducing the effect of environmental factors on an ANPR system. The ANPR system has been tested in multiple scenarios and is concluded to be sufficiently reliable.

## 5.2 Cost and Utility

1. Road and traffic security. To report over-speeding vehicles, absconding vehicles.

2. Law enforcement. For example: Border patrol, Road crime investigation, hit and run cases.

3. Parking Automation. Automatic parking saves time and ensures efficient management of parking space.

4. Access Control. Limiting access to certain areas where the road jamming and congestion are strictly prohibited. For example: Airports, government offices etc.

5. Journey time measurement.

6. ANPR Databases. Advanced ANPR technology keeps a database of detected car number plates. They have been very helpful in location tracking and activity history detailing during multiple crime investigations.

7. Motorway Road Tolling. This allows reduction of manual labor during tolling.

## 5.3 Project Links

1. Live working of the ANPR program:

https://drive.google.com/file/d/16FtmB4nf72bSE5sHBBZYtXO7vxrqaX_5/view

2. Project link. The code, datasets, images and videos are available on GitHub. We welcome recommendations and contributions to the project:

https://github.com/rajdeepadak/ANPR-using-Image-Processing-and-Deep-Learning

## 5.4 Scope for further work

Our team is still exploring more efficient ways to perform ANPR. Currently we are working on determining confidence level and predicting multiple outputs for one number plate. The data can also be uploaded to a remote server and we intend to use SQL for the same. We welcome the opinions of contributors.

# Bibliography

Digital Image Processing, Gonzalez and Woods, 3rd Edition, Pearson Publication.

Principles of Digital Image Processing Advanced Methods, Wilhelm Burger, Mark J. Burge, Springer.

Python Machine Learning by Example, Yuxi (Hayden) Liu, Packt Publications.

# Acknowledgements